\documentclass[conference]{IEEEtran}
\IEEEoverridecommandlockouts

\usepackage{booktabs}
\usepackage{cite}
\usepackage{amsmath,amssymb,amsfonts}
\usepackage{mathtools} 
\usepackage{algorithmic}
\usepackage{graphicx}
\usepackage{textcomp}
\usepackage{xcolor}
\usepackage{enumitem}          %
\usepackage{bm}

\usepackage{eso-pic}

\DeclareMathOperator*{\argmin}{arg\,min}

\def\BibTeX{{\rm B\kern-.05em{\sc i\kern-.025em b}\kern-.08em
    T\kern-.1667em\lower.7ex\hbox{E}\kern-.125emX}}
\begin{document}

\title{Relative Navigation and Dynamic Target Tracking for Autonomous Underwater Proximity Operations\\

}

\author{David Baxter$^{1,*}$, Aldo Terán Espinoza$^{2,*}$, Antonio Terán Espinoza$^{3}$, Amy Loutfi$^{1}$, \\ John Folkesson$^{4}$,
    Peter Sigray$^{2}$, Stephanie Lowry$^{1}$, and Jakob Kuttenkeuler$^{2}$%
    \thanks{*Equal contribution.}%
    \thanks{$^{1}$David Baxter, Stephanie Lowry, and Amy Loutfi are with the AI, Robotics and Cybersecurity Center, Örebro University, Örebro, Sweden (e-mail: \{david.baxter, stephanie.lowry, amy.loutfi\}@oru.se).}%
    \thanks{$^{2}$Aldo Terán Espinoza, Peter Sigray, and Jakob Kuttenkeuler are with the Centre for Naval Architecture, KTH Royal Institute of Technology, Stockholm, Sweden (e-mail: \{aldot, sigray, jakob\}@kth.se).}%
    \thanks{$^{3}$Antonio Terán Espinoza is with the department of Aeronautics and Astronautics, Massachusetts Institute of Technology, Cambridge, USA (e-mail: teran@mit.edu).}%
    \thanks{$^{4}$John Folkesson is with the Division of Robotics, Perception and Learning, KTH Royal Institute of Technology, Stockholm, Sweden (e-mail: johnf@kth.se).}%
    \thanks{This work was supported by SAAB and the Swedish Maritime Robotics Centre (SMaRC), and by the Wallenberg AI, Autonomous Systems and Software Program (WASP) funded by the Knut and Alice Wallenberg Foundation.}
}
\maketitle

\AddToShipoutPictureFG*{%
  \AtPageLowerLeft{%
    \raisebox{10mm}{\parbox{\paperwidth}{\centering
      \footnotesize This work has been submitted to the IEEE for possible publication. 
      Copyright may be transferred without notice, after which this version may no longer be accessible.
    }}%
  }%
}
\newcommand{\pcd}[1]{\prescript{#1}{}{\mathcal{P}}}
\newcommand{\point}[1]{\prescript{#1}{}{\bm{p}}}
\newcommand{\R}[0]{\mathbb{R}}  %
\newcommand{\I}[0]{\bm{I}}  %
\newcommand{\SE}[0]{\text{SE}}
\newcommand{\SO}[0]{\text{SO}}
\newcommand{\Wf}[0]{\text{W}}
\newcommand{\Tf}[0]{\text{T}}
\newcommand{\Cf}[0]{\text{C}}
\newcommand{\rotM}[0]{\bm{R}}
\newcommand{\Rot}[2]{\prescript{#1}{#2}{\mathbf{R}}}
\newcommand{\Tran}[3]{\prescript{#1}{}{\bm{t}_{#2/#3}}}
\newcommand{\TranHat}[3]{\prescript{#1}{}{\hat{\bm{t}}_{#2/#3}}}
\newcommand{\TranStar}[3]{\prescript{#1}{}{\bm{t}^*_{#2/#3}}}
\newcommand{\tran}[3]{\prescript{#1}{}{\mathbf{t}_{#2/#3}}}
\newcommand{\Unit}[3]{\prescript{#1}{}{\hat{\bm{u}}_{#2/#3}}}
\newcommand{\Vel}[3]{\prescript{#1}{}{\bm{v}_{#2/#3}}}
\newcommand{\vel}[3]{\prescript{#1}{}{\mathbf{v}_{#2/#3}}}
\newcommand{\Tfm}[2]{\prescript{#1}{#2}{\mathbf{T}}}
\newcommand{\TfmHat}[2]{\prescript{#1}{#2}{\hat{\mathbf{T}}}}
\newcommand{\TfmBar}[2]{\prescript{#1}{#2}{\bar{\mathbf{T}}}}
\newcommand{\TfmMat}[2]{\begin{bmatrix} #1 & #2 \\[0.3em] \bm{0}^\top & 1 \\[0.3em]\end{bmatrix}}
\newcommand{\Skew}[1]{{#1}^\wedge}
\newcommand{\RefCov}[1]{\prescript{#1}{}{\bm \Sigma}}
\newcommand{\RefEta}[1]{\prescript{#1}{}{\bm \eta}}
\newcommand{\p}[0]{\bm{p}}
\newcommand{\bmell}[0]{\bm{\ell}}
\newcommand{\z}[0]{\bm{z}}
\newcommand{\calX}[0]{\mathcal{X}}
\newcommand{\barA}[0]{\bar{A}}
\newcommand{\barB}[0]{\bar{B}}
\newcommand{\chx}[0]{\bm{x}^{\text{C}}}
\newcommand{\tgtx}[0]{\bm{x}^{\text{T}}}
\newcommand{\ttgtx}[0]{\bm{x}_{\text{TT}}}
\newcommand{\tgth}[1]{\prescript{W}{}{\theta}_{#1}}
\newcommand{\tgtv}[1]{\prescript{T}{}{v}_{#1}}
\newcommand{\ttgtv}[1]{\prescript{T}{}{\bm v}_{#1}}
\newcommand{\ttgtw}[1]{\prescript{T}{}{\bm \omega}_{#1}}
\newcommand{\imuf}[0]{\phi_{\text{imu}}}
\newcommand{\odomf}[0]{\phi_{\text{odom}}}
\newcommand{\rbf}[0]{\phi_{\text{rb}}}
\newcommand{\velf}[0]{\phi_{\text{v}}}
\newcommand{\depthf}[0]{\phi_{\text{d}}}
\newcommand{\mmf}[0]{\phi_{\text{mm}}}
\newcommand{\priorf}[0]{\phi_{\text{p}}}
\newcommand{\priorfc}[0]{\phi_{\text{p}_{\text{C}}}}
\newcommand{\priorft}[0]{\phi_{\text{p}_{\text{T}}}}
\newcommand{\expf}[0]{\phi_{\text{exp}}}
\newcommand{\redf}[0]{\phi_{\text{red}}}
\newcommand{\optf}[0]{\phi_{\text{opt}}}
\newcommand{\tof}[0]{\phi_{\text{to}}}

\newcommand{\calM}{\mathcal{M}}
\newcommand{\calZ}{\mathcal{Z}}
\newcommand{\Log}{\mathrm{Log}}
\newcommand{\Exp}{\mathrm{Exp}}
\newcommand{\Ad}{\mathrm{Ad}}
\newcommand{\Jr}{J_r}
\newcommand{\Jl}{J_l}

\newcommand{\ad}{\mathrm{ad}}
\newcommand{\se}{\mathfrak{se}}
\newcommand{\bth}{\boldsymbol{\theta}}   %
\newcommand{\bfrho}{\boldsymbol{\rho}}   %

\newcommand{\bfI}{\mathbf{I}}
\newcommand{\bfzero}{\mathbf{0}}
\newcommand{\bfQ}{\mathbf{Q}}

\newcommand{\AdM}[1]{\Ad_{#1}}           %
\newcommand{\mjac}[2]{J_{#2}^{#1}}
\newcommand{\Jlinv}[1]{J_l^{-1}(#1)}
\newcommand{\Jrinv}[1]{J_r^{-1}(#1)}

\newcommand{\chxinv}{{\bm{x}^{\text{C}}}^{-1}}
\begin{abstract}
    Estimating a target’s 6-DoF motion in underwater proximity operations is difficult because the chaser lacks target-side proprioception and the available relative observations are sparse, noisy, and often partial (e.g., Ultra-Short Baseline (USBL) positions). Without a motion prior, factor-graph maximum a posteriori estimation is underconstrained: consecutive target states are weakly linked and orientation can drift.
    We propose a generalized constant-twist motion prior defined on the tangent space of Lie groups that enforces temporally consistent trajectories across all degrees of freedom; in SE(3) it couples translation and rotation in the body frame. We present a ternary factor and derive its closed-form Jacobians based on standard Lie group operations, enabling drop-in use for trajectories on arbitrary Lie groups.
    We evaluate two deployment modes: (A) an SE(3)-only representation that regularizes orientation even when only position is measured, and (B) a mode with boundary factors that switches the target representation between SE(3) and 3D position while applying the same generalized constant-twist prior across representation changes.
    Validation on a real-world dynamic docking scenario dataset shows consistent ego-target trajectory estimation through USBL-only and optical relative measurement segments with an improved relative tracking accuracy compared to the noisy measurements to the target. Because the construction relies on standard Lie group primitives, it is portable across state manifolds and sensing modalities.
\end{abstract}

\begin{IEEEkeywords}
    target tracking, state estimation, factor graphs, motion priors, proximity operations, autonomous docking, autonomous underwater vehicle navigation.\end{IEEEkeywords}
\section{Introduction and Motivation}
Estimating a target’s six-degree-of-freedom (6-DoF) motion from sparse and partial measurements is a fundamental challenge in nonlinear state estimation. Without additional structure, the inference problem is underdetermined and trajectory estimates quickly become physically inconsistent.
This leads to a central question:

\textit{How can we model and enforce physically consistent motion on a nonlinear state manifold when measurements are intermittent and underconstraining?}

\begin{figure}[t]
  \centering
  \includegraphics[width=0.85\linewidth]{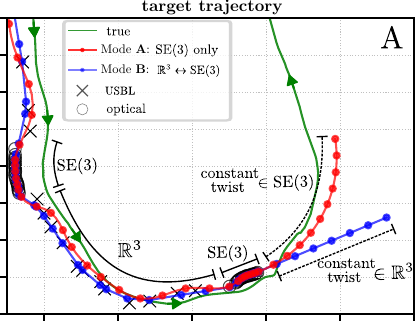}
  \caption{
    Zoom-in on Region~\textbf{A} of Fig.~\ref{fig:full_trajectories}, showing incremental target estimates for Mode A and Mode B. Solid black bracketed labels indicate the manifold of the relative measurement type constraining each trajectory segment: measurement updates (USBL in $\R^3$, optical in $\SE(3)$). Although Mode~A maintains an $\SE(3)$ representation throughout, these labels also correspond to the manifold used to represent the target by Mode~B along the same portions of trajectory. At the far right, where no measurements are available, both methods extrapolate forward: Mode~A continues with a constant-twist prior in $\SE(3)$, while Mode~B continues with a constant-twist prior in $\R^3$. Here the dotted black lines represent the manifold of the constant twist prior.
  }
  \label{fig:moneyplot}
\end{figure}

We address this challenge by introducing a constant-twist motion prior that enforces temporally consistent trajectories directly on the tangent space of the Lie group representing the state. This prior acts as a physically meaningful regularizer, allowing robust estimation even when measurements are intermittent and underconstraining.
Although our evaluation centers on underwater proximity operations, the proposed prior is domain-agnostic and applicable to any robotic setting where physically consistent motion must be inferred under partial observability.

This challenge is particularly pressing in underwater robotics, where Autonomous Underwater Vehicles (AUVs) are frequently used for a wide range of underwater operations throughout a large span of different domains: from environmental monitoring to close inspection and manipulation~\cite{jacobi_inspection_2015}. However, because of the hostile operational environment and challenging sensor modalities, the development of methods and techniques aimed for AUVs to handle increasingly complex tasks lags behind their robotic counterparts on land and air~\cite{gonzalez_survey_2020}. Some of the most complex tasks for a robot to handle are those that require their interaction or collaboration in close proximity with other agents or elements in the environment. Underwater operations such as docking, manipulation of intervention panels, ship hull inspection, among others, all fall under this category of proximity operations---\textit{prox-ops} for short.

Underwater prox-ops follow a set scheme where a chasing agent handles a series of consecutive phases to reach a target agent in order to execute a joint maneuver, for example: inspection, manipulation, or docking. These phases are dependent on the distance to the target given the specific sensor suite available for acquiring relative measurements: the long distance phase typically uses only acoustic measurements since only acoustic signals can travel large distances underwater, whereas optical or electromagnetic measurements are commonly used at terminal ranges. The present work addresses the specific scenario of a chaser AUV homing to a dynamically active target---a prerequisite for several more specific prox-ops that has not been researched thoroughly~\cite{teran_simultaneous_2023}.
The initial task of the autonomous chaser during a dynamic homing scenario is that of estimating both its own trajectory and the trajectory of the dynamic target. This estimation must be performed using sparse, noisy, aperiodic, and often asymmetric measurements that provide only partial observability of the target’s state, with varying levels of noise in the measured dimensions.
For example, observations may consist of pure bearing, which lies on the unit sphere $S^2$, or range–bearing, which lies on the product manifold $\mathbb{R}^+ \times S^2$ with a three-dimensional tangent space locally equivalent to $\mathbb{R}^3$. In an arbitrary probabilistic state estimation approach, the required measurement model may only constrain certain components of the state and a subset of degrees of freedom can remain underconstrained, preventing the estimates from converging to their true values.
Thus, we must fuse sparse, underconstraining measurements with motion priors that regularize unobservable degrees of freedom to ensure physically plausible trajectories.
The main purpose of this work is to develop a general, minimally restrictive solution to dynamic target tracking, which is naturally formulated as a nonlinear state estimation problem. To this end, we introduce a constant-twist motion prior that regularizes trajectory evolution without reliance on application-specific design, constraining otherwise unobservable degrees of freedom and enabling robust estimation even under sparse and partial measurements. While our experiments validate the approach on an underwater proximity-operations dataset, the construction uses only standard Lie group primitives $(\Exp/\Log,\, \oplus/\ominus,\, \Ad,\, \Jl,\, \Jr)$ and is therefore portable across state manifolds and sensing modalities.

\noindent
\textbf{The key contributions of this work are:}
\begin{itemize}
  \item We formulate a general \textit{constant-twist motion prior} that operates directly on the tangent space of any Lie group manifold, enforcing temporally consistent motion in a principled and reusable way, and derive a custom factor implementing this prior.
  \item We integrate this factor into a factor graph-based state estimation framework and demonstrate its effectiveness for joint AUV and target trajectory estimation in a real-world dynamics rendezvous scenario. We show that the factor seamlessly adapts to changes in representations, effectively handling both pure $\SE(3)$ and mixed $\R^3~\leftrightarrow~\SE(3)$ states.
\end{itemize}
The present work reads as follows: the following section (Section~\ref{sec:preliminaries}) will set the scene for the related work and the developed methods by laying out the background theory and the problem formulation used to answer our research question; Section~\ref{sec:related_work} details the state of the art on (1) target tracking for dynamic underwater prox-ops in general, and (2) similar methods that address the subject of smooth motion priors for factor graph-based state estimation; the inner workings of our main contributions are addressed in Section~\ref{sec:methods} and its implementation and evaluation using a real-world dataset is included in Section~\ref{sec:evaluation}. We conclude this work with a discussion regarding the limitations and future work in Section~\ref{sec:conclusions}.
\section{Preliminaries and Problem formulation}
\label{sec:preliminaries}
\subsection{Problem Setup}
\label{subsec:setup}
\begin{figure}[t]
    \centering
    \includegraphics[width=\linewidth]{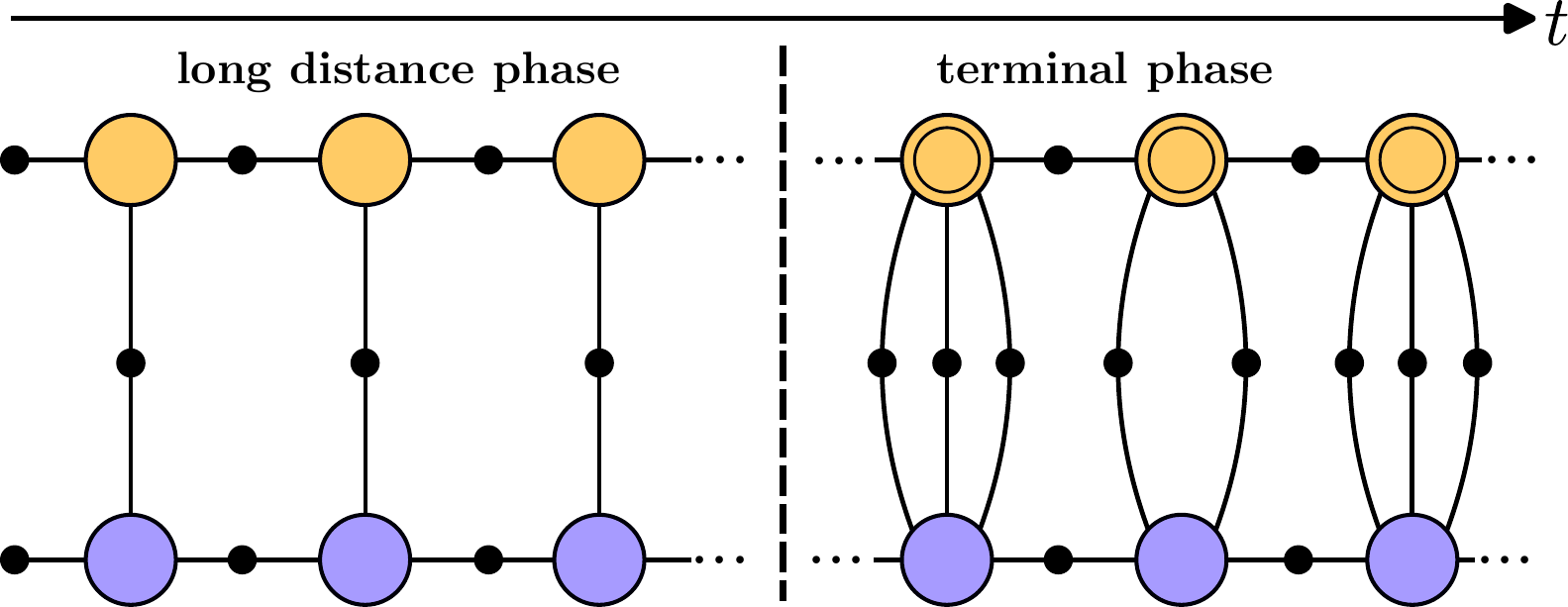}
    \caption{
        Factor graph modeling the long distance and terminal navigation phases the chaser agent must handle to reach its target. The chained, blue and orange nodes represent the evolution of the chaser's and target's state through time, respectively. The edges connecting both chains together represent the relative measurements acquired through the different phases. The difference in the orange nodes (single-ring vs double-ring circle) between phases suggest a change of the target's state representation.
    }
    \label{fig:proxops_graph}
\end{figure}

We consider a relative–navigation scenario in which an autonomous
vehicle ---the \emph{chaser}---must track a dynamic
\emph{target} while simultaneously localizing itself in preparation for
close proximity operations.
At discrete times $t_k$ $(k=0,\dots,N)$ we denote the chaser and target states by
$\chx_k\in\calM$ and $\tgtx_k\in\calM$ respectively, where
$\calM$ is a smooth, connected manifold that describes rigid‐body
configuration.
Common choices include the Lie‐groups $\SE(2)$ for planar motion,
$\SO(3)$ for attitude‐only problems, and~$\SE(3)$ for full 6‐DoF motion.
The platform receives aperiodic measurement sets
\mbox{$\calZ=\{\!z_i\!\}$} comprising
\begin{itemize}
    \item \textbf{ego–motion factors} (e.g.\ odometry, Doppler logs);
    \item \textbf{partial relative measurements} such as range, bearing, or partial
          $\R^3$ offsets $(x,y,z)$ to the target; and
    \item \textbf{intermittent relative full‐pose observations} that may only become
          available at close range as in our applied example.
\end{itemize}
These measurements are inherently \emph{asymmetric}: they constrain some state components (e.g., position or range) much more strongly than others (e.g., yaw), often leaving parts of the target pose effectively unobserved. Robust inference therefore requires motion priors that supply the missing structure for these unobservable degrees of freedom.
We model this information fusion problem in a probabilistic inference framework which leverages factor graphs as a modular and generalizable representation for the interdependence between the latent variables (agent's trajectories) and the acquired relative measurements~\cite{dellaert_factor_2017}. As shown in~\cite{teran_simultaneous_2023}, factor graphs present a versatile solution for arbitrary prox-ops scenarios by intuitively tracking the agent's trajectories in separate variable chains that are tied together through the hitherto described relative measurements. Fig.~\ref{fig:proxops_graph} illustrates the underlying factor graph of a general prox-ops scenario: two variable node chains (top for target and bottom for chaser) are joined together through relative measurements. The evolution of their state is constrained via motion model predictions (factors between target states) or odometry measurements (factors between consecutive chaser state). Given that during the terminal phase it is possible for the chaser to switch sensor modality to be able to observe the full state of the target, the changes in the variable nodes for the target suggest the possibility of a change of representation between phases---although not necessary as shown later.

We cast trajectory estimation as Maximum A Posteriori (MAP) inference in the described nonlinear factor graph~\cite{dellaert_factor_2017}, expressed as
\begin{equation}
    \hat{\mathbf{x}}
    =\argmin_{\mathbf{x}}
    \sum_{i}\,\phi_i\bigl(\mathbf{x}_{\mathcal{S}(i)}\bigr)\; ,
    \label{eq:MAP}
\end{equation}
where $\hat{\mathbf{x}}$ collects all states
$\{\chx_k,\tgtx_k\}_{k=0}^{N}$,
each factor $\phi_i$ encodes the negative log of a probabilistic relation,
and $\mathcal{S}(i)$ denotes the subset of variables touched by factor~$i$.
The formulation of the estimation problems follows: let $\Phi_\text{meas}$ denote the set of measurement factors derived from
$\calZ$ and $\Phi_\text{motion}$ the set of motion priors.
The overall MAP objective is
\begin{equation}
    \hat{\mathbf{x}}
    = \argmin_{\mathbf{x}}
    \sum_{\phi \in \Phi_\text{meas}}
    \phi(\mathbf{x}_{\mathcal{S}(\phi)})
    + \sum_{\psi \in \Phi_\text{motion}}
    \psi(\mathbf{x}_{\mathcal{S}(\psi)}),
    \label{eq:fg_cost}
\end{equation}
where $\|\bm\epsilon\|_{\Sigma}^2=\bm\epsilon^{\!\top}\Sigma^{-1}\bm\epsilon$
denotes the Mahalanobis distance of residual~$\bm\epsilon$.
In practice, a factor graph for joint ego–target estimation will also
include additional priors (e.g., initial state configurations or attitude
priors). For compactness, we subsume these into $\Phi_\text{meas}$ in
the formulation above.

Optimization proceeds in the tangent space of the state manifold~$\calM$.
Each state $x \in \calM$ is updated using a local coordinate increment
$\delta \in T_{x}\calM \cong \R^{\dim\calM}$ together with a
\emph{retraction}\,%
\footnote{When $\calM$ is a Lie‐group, robotics libraries often choose the exponential map
    $\mathrm{Exp}\colon T_{x}\calM\rightarrow\calM$, but any first‐order
    retraction suffices. GTSAM~\cite{gtsam} implements this abstraction through methods such as
    \texttt{retract} and \texttt{localCoordinates}; it also supplies analytic
    Jacobians for core Lie‐group operations.}
\begin{equation}
    \mathsf{Ret}_x(\delta)\;:\;T_{x}\calM\to\calM.
    \label{eq:retract}
\end{equation}
By performing this optimization in the tangent space, the underlying state variables which potentially lie on a non-linear manifold can be optimized in a linear vector space where it is comparatively simple to take derivatives and integrals over the evolution of these states~\cite{Sola2018MicroLie}.
\subsection{Motion Priors via Body‐Frame Twist}
\label{subsec:motion_priors}

Traditional smoothness terms for target tracking often assume constant velocity in~$\R^n$. While simple, such Euclidean priors fail to precisely model rotational dynamics and degrade in performance or fail to converge when the measurement set undersamples orientation.

A more principled alternative is to express motion as a constant \textbf{body-frame twist}\footnote{In robotics and screw theory, a \emph{twist} denotes an element of the Lie algebra $\mathfrak{se}(3)$ (or $\mathfrak{se}(2)$) representing instantaneous rigid-body velocity. More generally, the same construction applies to any smooth matrix Lie group $G$, in which case ``twist'' should be read as the corresponding Lie algebra element $\boldsymbol{\xi} \in \mathfrak{g}$. Throughout this work, we use $\delta \in \mathfrak{g}$ to denote a \emph{tangent-space increment} (units of m, rad for $\mathfrak{se}(3)$) and $\boldsymbol{\xi} \in \mathfrak{g}$ to denote a \emph{tangent-space velocity} (units of m/s, rad/s for $\mathfrak{se}(3)$), related by $\delta = \boldsymbol{\xi} \, \Delta t$.}
, represented as a vector $\bm\xi \in \mathfrak{m}$ in the tangent space associated with the manifold $\calM$ representing the state, where $\bm\xi$ is the rate of change of the state variable in $\calM$ as mapped to the variable's tangent space. Over a period of time t, the resulting displacement increment in the variable's tangent space $\bm\delta = \bm\xi \Delta t$.

This formulation naturally couples translational and rotational components when $\calM$ is a matrix Lie group with those components, and allows motion priors to be expressed as smooth trajectories in the local tangent space.

In the case of rigid-body motion in three dimensions, where $\calM = \SE(3)$, the twist $\bm\xi \in \mathfrak{se}(3)$ is a six-dimensional vector that combines angular and linear velocity in the body frame. We use this structure in our implementation, but the approach applies equally to other manifolds (e.g., $\SE(2)$, $\SO(3)$, $\SE_2(3)$) with appropriate definitions for the twist, retraction, and local coordinates.

\paragraph{Constant‐Twist Hypothesis}
Assuming the body‐frame twist is \emph{piecewise constant} over two
consecutive sampling intervals, it produces a first‐order, physically
consistent kinematic model that: (i) couples three consecutive poses through a shared twist estimate, (ii) preserves the sparsity of the factor graph by involving only a minimal ternary factor, and (iii) provides regularization across all degrees of freedom, without imposing hard constraints.

\paragraph*{Focus of this work}
Our focus here is on a \emph{constant-twist motion prior}, which defines
the motion model in our factor graph and provides a physically
meaningful, Lie group–agnostic regularizer. In the next section, we present related works before returning to our constant twist prior in
Sec.~\ref{sec:methods} where we describe its general construction and then
instantiate it on $\SE(3)$ and $\R^3$ with closed-form Jacobians. Complete
derivations are provided in the Appendix.
\section{Related Work}
\label{sec:related_work}

The state of the art on autonomous underwater prox-ops is mostly represented by works that try to solve an instance of the underwater docking problem, with most works only tackling scenarios with passive targets---the survey detailed in~\cite{teran_simultaneous_2023} shows that only a small percentage of work deal with a dynamic target. To further narrow our technical gap, we rule out methods that solve the dynamic target tracking problem by means of guidance methods, i.e., methods that couple perception cues directly to the chaser's control system to track the target in a reactive manner; these methods suffer from the fact that the chaser does not explicitly reason about the dynamics of the target which are essential for more complex rendezvous/homing maneuvers.

In previous work~\cite{teran_simultaneous_2023}, where the long distance homing scenario with noisy and sparse Ultra-short Baseline (USBL) measurements is solved using a factor graph framework, the target’s motion was simplified by assuming a globally constant motion prior on the heading and velocity, and the assumed dynamics effectively only update the target's position in~$\mathbb{R}^2$. While effective for that setting, this formulation relied on application-specific assumptions that limit generality. In the present work, we instead propose a Lie group prior that captures both translation and rotation in a general and geometrically consistent manner, enabling application across domains.
To maintain a fully constrained problem as sensing modalities change across prox-op phases, \cite{espinoza2024boundaryfactors} introduced boundary factors, which transfer information between changing target representations. Specifically, the target is represented in~$\mathbb{R}^3$ when only range--bearing measurements are available, and upgraded to~$\SE(3)$ when full relative pose is observed, with boundary factors ensuring consistency across transitions.
The authors in~\cite{ruan_factor_2020} addressed dynamic docking with a constant-velocity prior on the target’s~$\mathbb{R}^3$ position. Our approach instead computes velocities implicitly from pose differences and generalizes the prior to arbitrary manifolds. In~\cite{yu_dynamic_2025}, they avoided target dynamics entirely by formulating docking as a relative localization problem; while lightweight, this loses the predictive power of a motion model to constrain unobservable degrees of freedom.
In~\cite{STEAM, Full_STEAM}, Gaussian Process regression was used to model continuous robot trajectories, with factor graph variables treated as sparse GP support states. Although their primary intent was to estimate continuous trajectories using a sparse set of support states amenable to factor graph based optimization, they formulate a prior on the state that is similar in effect to the one we present here. This formulation was extended to factor graph optimization with landmark measurements (STEAM), and later to motion planning~\cite{Mukadam-IJRR-18}.
The closest related approach is~\cite{dong2018sparse}, who formulated a constant-velocity GP prior directly on the Lie algebra of the state. Unlike their method, our factor does not require maintaining derivative states as additional graph variables. We also provide a practical recipe and analytical Jacobians for implementing the constant-twist factor on any Lie group, and demonstrate it on both~$\SE(3)$ and~$\mathbb{R}^3$, and demonstrate their application in a tracking problem with and without a change in the target’s state representation throughout the trajectory.
In the subject of pure target tracking, \cite{petersen_integrated_2023} formalizes the notion of a constant velocity motion model for connected and unimodular Lie groups and showcase its performance with an integrated probabilistic data association filter in a ground vehicle tracking scenario in $\SE(2)$.

We do not claim to be the first to jointly estimate a chaser and a moving target in a coupled factor graph from the chaser’s perspective—prior work has done so while modeling the target’s dynamics in the Euclidean space~$\mathbb{R}^3$ with a constant-velocity prior. Our contribution with respect to these works, is to replace that Euclidean dynamics factor with a \emph{general} Lie group, constant-twist prior and integrate it into the same coupled estimation framework. This formulation models both translation and rotation in a geometrically consistent way, uses $\oplus/\ominus$ operations and adjoint/Jacobian structure for correct frame handling, and naturally reduces to~$\mathbb{R}^n$ when only positions are needed. Practically, it enables curved, fully 6-DoF target trajectories to be constrained by a body-frame velocity prior, improves consistency when states are partially observed, and supports clean transitions between~$\SE(3)$ and~$\mathbb{R}^3$ via boundary factors when measurement modalities change.
\section{Constant Twist Motion Prior}
\label{sec:methods}

We introduce a motion prior based on the assumption of constant body-frame twist across consecutive time intervals. This factor provides a physically meaningful constraint that can regularize both translation and rotation, making it particularly useful when the state is only partially observed. Although our implementation in Section \ref{sec:evaluation} targets $\SE(3)$, the derivation in \ref{sec:residual_definition} is general and applies to any manifold, including the Lie groups commonly used in robotics applications.

\paragraph{Notation and Conventions.}
We denote $\delta \in \mathfrak{g}$ a \emph{tangent-space increment} and $\boldsymbol{\xi} \in \mathfrak{g}$ a \emph{tangent-space velocity}, related via $\delta = \boldsymbol{\xi} \, \Delta t$. For $\mathfrak{se}(3)$, the translation components have units of meters and the rotation components have units of radians; for other manifolds, the corresponding natural units apply. The operators $\Log$ and $\Exp$ denote the Lie group logarithm and exponential maps, respectively, mapping between the manifold $\mathcal{M}$ and its Lie algebra $\mathfrak{g}$. Group composition is written $T_a T_b$ and inversion as $T^{-1}$.

We adopt the common $\oplus$ and $\ominus$ notation for manifold operations
\cite{Sola2018MicroLie,HERTZBERG201357}. For a point $X \in \mathcal{M}$ on
a smooth manifold and a tangent vector $\delta \in T_X \mathcal{M}$, the \emph{plus} operator
\[
  X \oplus \delta \ \in \ \mathcal{M}
\]
denotes the \emph{retraction} of $\delta$ onto the manifold at $X$.
The \emph{minus} operator
\[
  Y \ominus X \ \in \ T_X \mathcal{M}
\]
maps two points $X, Y \in \mathcal{M}$ to the tangent space at $X$,
yielding the local coordinates of $Y$ relative to $X$.
For matrix Lie groups, these reduce to
$X \oplus \delta = X \Exp(\delta)$ and $Y \ominus X = \Log(X^{-1} Y)$.

\subsection{Overview}

The constant-twist motion prior works by estimating the body-frame twist from two consecutive poses and projecting it forward to predict the third. The residual is defined as the deviation between this prediction and the actual third pose, evaluated in the tangent space of the manifold from the frame of reference of the projected pose.

Let $T_{k-1}, T_k, T_{k+1} \in \calM$ be three consecutive poses. Let $\Delta t_1 = t_k - t_{k-1}$ and $\Delta t_2 = t_{k+1} - t_k$. The factor enforces that the body-frame twist inferred from $(T_{k-1}, T_k)$ remains consistent with the displacement from $T_k$ to $T_{k+1}$.

\begin{figure}[t]
  \centering
  \includegraphics[width=0.65\linewidth]{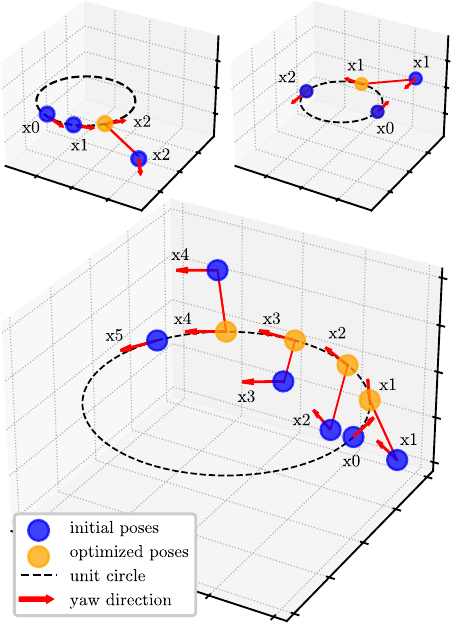}
  \caption{Illustration of the ternary constant-twist SE(3) factor.
    \textbf{(Top-left)} Two poses are anchored with priors on the unit circle; a third pose is noisily initialized.
    The ternary constant-twist factor (applied to the triplet) predicts the future pose and snaps it to the same circular arc implied by the anchored poses.
    \textbf{(Top-right)} First and last poses are anchored; the factor constrains the intermediary pose to lie on the consistent arc (interpolation between measurements).
    \textbf{(Bottom)} Two endpoint poses ($x_0,x_5$) are anchored; four intermediary poses ($x_1\!-\!x_4$) are noisily initialized.
    Chaining the constant-twist factor sequentially yields a minimum-cost configuration placing all intermediates along the same arc.
    Only priors on the anchored endpoints and the constant-twist factors are used; the circular arc emerges from the motion model, not from an explicit geometric constraint.}  \label{fig:unit_circ_test}
\end{figure}

\subsection{Residual Definition}
\label{sec:residual_definition}
The residual $\bm\epsilon_k$ is computed in five steps:
\begin{enumerate}[label=(\roman*)]
  \item \textbf{Relative increment:}
        \begin{equation}
          \delta_1 = T_k \ominus T_{k-1} \ \in \ T_{T_{k-1}}\mathcal{M}
          \label{eq:delta1}
        \end{equation}
        \quad On Lie groups, this reduces to
        $\delta_1 = \Log(T_{k-1}^{-1} T_k)$.

  \item \textbf{Body–frame velocity estimate:}
        \begin{equation}
          \hat{\boldsymbol{\xi}} = \frac{\delta_1}{\Delta t_1}
          \label{eq:xi_hat}
        \end{equation}

  \item \textbf{Predicted next increment:}
        \begin{equation}
          \delta_2 = \hat{\boldsymbol{\xi}} \Delta t_2
          \label{eq:delta2}
        \end{equation}

  \item \textbf{Predicted next pose:}
        \begin{equation}
          \hat{T}_{k+1} = T_k \oplus \delta_2
          \label{eq:pred_model}
        \end{equation}
        \quad On Lie groups,
        $\hat{T}_{k+1} = T_k \Exp(\delta_2)$.

  \item \textbf{Residual:}
        \begin{equation}
          \bm\epsilon_k = T_{k+1} \ominus \hat{T}_{k+1} \ \in \ T_{\hat{T}_{k+  1}}\mathcal{M}
          \label{eq:residual_ct}
        \end{equation}
        \quad On Lie groups,
        $\bm\epsilon_k = \Log(\hat{T}_{k+1}^{-1} T_{k+1})$.
\end{enumerate}

This residual penalizes deviations from constant body-frame velocity (twist), acting as a smoothness prior across the sequence. We derive the full analytical Jacobians of the residual $\bm\epsilon_k$ with respect to all three poses:
\[
  \frac{\partial \bm\epsilon_k}{\partial T_{k-1}}, \quad
  \frac{\partial \bm\epsilon_k}{\partial T_k}, \quad
  \frac{\partial \bm\epsilon_k}{\partial T_{k+1}}.
\]

These are computed via the chain rule, linking the intermediate operations in the residual calculation. We provide the full derivation in the Appendix.

Figure~\ref{fig:unit_circ_test} illustrates the effect of the constant-twist prior through several examples in $SE(3)$. The factor is applied to a sequence of poses initialized on and around a unit circle embedded in a 3-D space, demonstrating how the prior extrapolates the newest pose and regularizes intermediate poses.

\subsection{Extending to Other Manifolds}
The five–step residual in \eqref{eq:delta1}–\eqref{eq:residual_ct} is group-agnostic. To use it on any Lie group $\calM^\star$, it suffices to define (i) a minimal coordinate map for the Lie algebra $\mathfrak{m}^\star$ with hat/vee,
\footnote{In the Appendix, we follow the convention of \cite{Sola2018MicroLie} for $\se(3)$ and stack translation then rotation, $\boldsymbol{\xi}=\begin{bmatrix}\bm\rho\\ \bm\theta\end{bmatrix}\!\in\!\mathbb{R}^6$, with $\bm\rho,\bm\theta\in\mathbb{R}^3$. Some libraries (e.g., GTSAM’s \texttt{Pose3}) use the opposite ordering, $\boldsymbol{\xi}_{\text{GTSAM}}=\begin{bmatrix}\bm\theta\\ \bm\rho\end{bmatrix}$. The hat/vee maps respect the chosen ordering; for the $\se(3)$ convention in our Appendix, $\boldsymbol{\xi}^\wedge=\begin{bmatrix}[\bm\theta]_\times & \bm\rho\\ \mathbf{0}^\top & 0\end{bmatrix}$ and $(\cdot)^\vee$ extracts $(\bm\rho,\bm\theta)$.}
(ii) a retraction via $\Exp/\Log$ (equivalently $\oplus/\ominus$), (iii) the group product and inverse, and (iv) the standard differentials—adjoint $\Ad_X$ and left/right Jacobians $J_l, J_r$ (and their inverses). With these in place, the pipeline $\delta_1=\Log(T_{k-1}^{-1}T_k)$, $\hat{\boldsymbol{\xi}}=\delta_1/\Delta t_1$, $\delta_2=\hat{\boldsymbol{\xi}}\Delta t_2$, $\hat T_{k+1}=T_k\Exp(\delta_2)$, and $\bm\epsilon_k=\Log(\hat T_{k+1}^{-1}T_{k+1})$ applies verbatim; only the group-specific operators change.
No structural changes to the factor graph structure are required. This modularity allows reuse across different robotic systems with minimal adaptation.
\section{Implementation and Evaluation}
\label{sec:evaluation}
\begin{figure}[t]
    \centering
    \includegraphics[width=\linewidth]{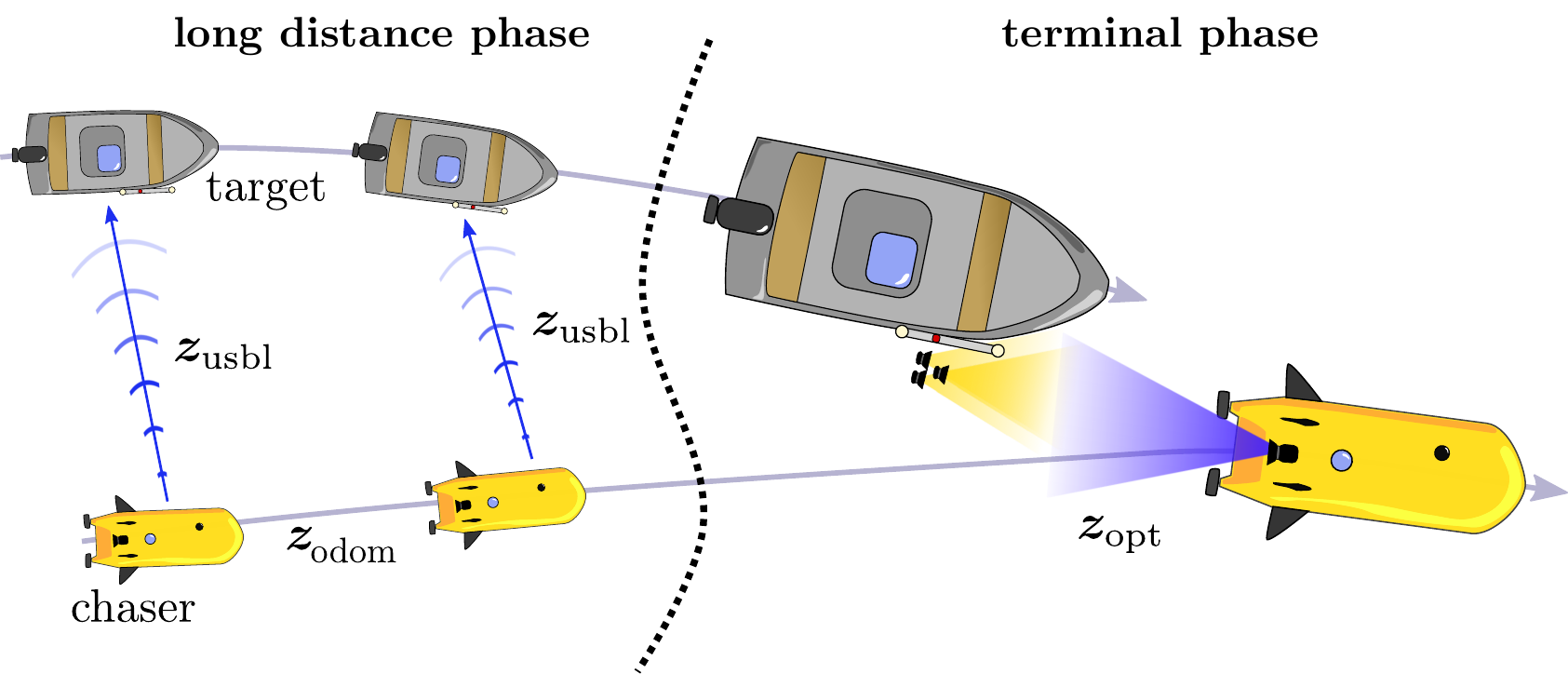}
    \caption{Pictorial description of the dynamic homing scenario used for the implementation and evaluation of our methods. The chaser AUV equipped with a USBL positioning system and a monocular camera, captures relative measurements to a surface vessel at the different phases of the operation. Figure adapted from~\cite{teran_consistent_2025}.}
    \label{fig:asko_scenario}
\end{figure}

\begin{figure}[t]
    \centering
    \includegraphics[width=\linewidth]{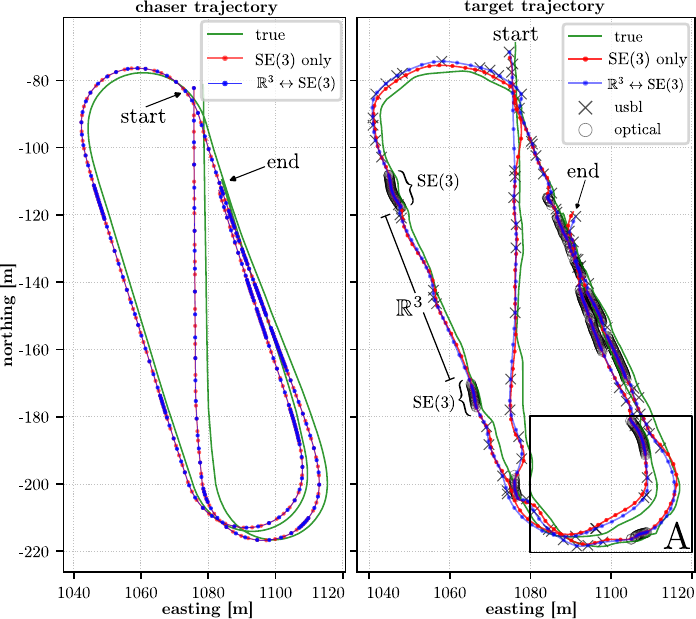}
    \caption{Resulting estimated and ground truth trajectories for both the chaser and the target. The $\SE(3)$ and $\R^3$ annotations on the target's plot detail the region's sensor modality. The delimited region \textbf{A} shows the fully smoothed solution of the transect shown in Figure~\ref{fig:moneyplot} after obtaining more measurements.}
    \label{fig:full_trajectories}
\end{figure}

To experimentally validate the performance of the constant-twist prior, we make use of the open datasets curated in~\cite{teran_consistent_2025}. The dataset's scenario, pictorially described in Fig.~\ref{fig:asko_scenario}, consists of a dynamic opto-acoustic rendezvous between a surface vessel (the target agent) and an AUV (chaser agent). The AUV follows a predetermined trajectory underwater (shown in Fig.~\ref{fig:full_trajectories}, left) and the surface vessel follows at diverse distances to simulate both the long-distance and terminal phases of the proximity operation. During the long distance phase, the chaser $\Cf$ measures only a relative position to the target $\bm z_{\text{usbl}} = \Tran{\Cf}{\Tf}{\Cf}$ through a USBL localization system, whereas during the terminal phase, additional high-accuracy, full 6-DoF optical relative measurements $\bm z_{\text{opt}} = \Tfm{\Cf}{\Tf}$ are computed using a monocular camera on the chaser and a set of light fiducials on the target.
\footnote{For translation vectors we make a third index explicit to denote the expression frame.
    For example, $\Tran{\Cf}{\Tf}{\Cf}$ is the displacement from $\Cf$ to $\Tf$ expressed in the chaser frame $\Cf$,
    while the \emph{same geometric displacement} expressed in the world frame is
    $\Tran{\Wf}{\Tf}{\Cf} = \Rot{\Wf}{\Cf}\,\Tran{\Cf}{\Tf}{\Cf}$.}
The chaser's odometry $\bm z_{\text{odom}} = \Tfm{\Cf_{k-1}}{\Cf_{k}}$ is measured by an onboard navigation solution. The dataset provides both the raw measurements and ground truth for both of the agents' trajectories.
\subsection{Factor graph implementation}
\label{sec:fg_methods}
\begin{figure}[t]
    \centering
    \includegraphics[width=\linewidth]{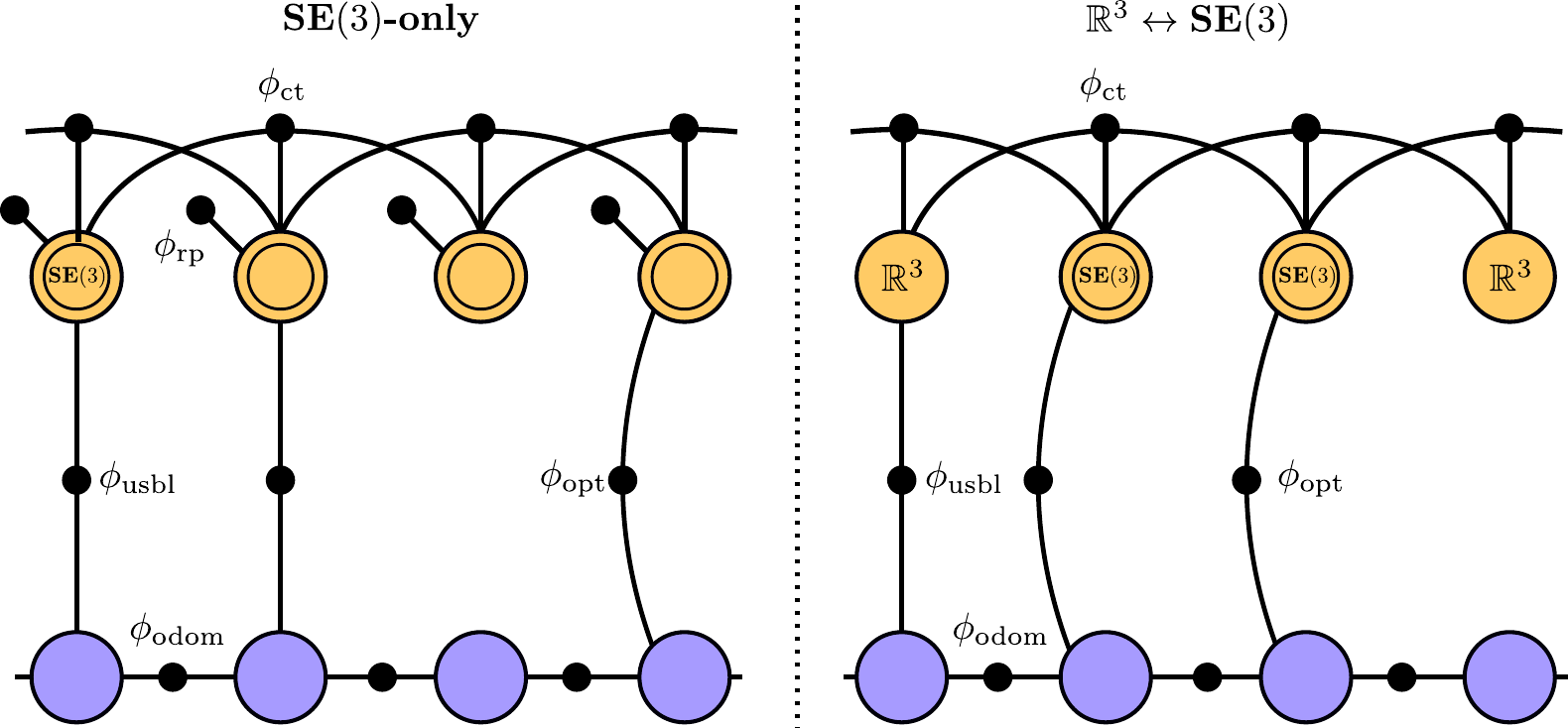}
    \caption{Two proposed approaches for the implementation of the constant twist factor ($\phi_{\text{ct}}$) posed as a factor graph relative trajectory estimation problem. \textbf{(Left)} Mode A: An $\SE(3)$ only representation is kept for the target's state chain and $\phi_{\text{ct}}$ constrains and predicts its motion throughout the different operation regimes: fully constrained, partially constrained, and unconstrained by relative measurements. \textbf{(Right)} Mode B: The target state chain consists of both $\R^3$ and $\SE(3)$ states that switch depending on the measurement type while $\phi_{\text{ct}}$ constrains and predicts its evolution on the appropriate manifold.
    }
    \label{fig:graphs}
\end{figure}

We evaluate the constant–twist prior in two factor–graph configurations that differ in their \emph{target–state representation policy}. Figure~\ref{fig:graphs} illustrates both modes.

\paragraph{Mode A ($\SE(3)$ only)}
In this formulation, the target state chain is represented entirely in $\SE(3)$, with consecutive states connected by the constant-twist factor $\phi_{\text{ct}}$. This prior constrains both translation and orientation, providing a smooth and physically consistent trajectory even in intervals where orientation is not directly observed. While orientation uncertainty grows in the absence of measurements, the trajectory remains defined.

To further encode domain knowledge about the target vessel, we add a soft
roll--pitch prior factor $\phi_{\text{rp}}$ in Mode~A. This factor penalizes
departures from an upright orientation by measuring the difference between the
target's body frame and a reference world frame yawed to match the vessel's
heading, expressed in the Lie algebra of $\SE(3)$. Without this additional
constraint, the optimizer can still find mathematically feasible
trajectories---including 180$^\circ$ rolls or other inverted motions---that are
consistent with USBL-only position measurements but not physically realistic
for a surface vessel. The roll--pitch prior therefore demonstrates how
application-specific knowledge can be incorporated as a soft factor to avoid
unphysical modes, without undermining the generality of the constant--twist
framework.

\paragraph{Mode B ($\R^3 \leftrightarrow \SE(3)$)}
In this formulation, we follow the boundary factors framework of~\cite{espinoza2024boundaryfactors} to enable automatic switching between representations: $\SE(3)$ when optical measurements are available, and $\R^3$ when only USBL measurements constrain the target. Crucially, the same constant-twist factor $\phi_{\text{ct}}$ is applied throughout, instantiated in the appropriate manifold. In $\R^3$, it reduces to constraining translational motion only, while in $\SE(3)$ it constrains both translation and orientation. Even without using $\phi_{\text{ct}}$, this ensures that the graph remains consistently constrained across modalities, with the representation of the state adapting to the information content of the available measurements.

Both graphs share the same general MAP objective in Eq.~\eqref{eq:fg_cost}, populated by motion factors $\Phi_{\text{motion}}$ and measurement factors $\Phi_{\text{meas}}$.
\paragraph{Constant–twist factor}
The constant–twist factor takes the form
\begin{equation}
    \phi_{\text{ct}}(\tgtx_{k-1}, \tgtx_k, \tgtx_{k+1})
    = \bigl\|\,\bm\epsilon_{\text{ct}}(\tgtx_{k-1}, \tgtx_k, \tgtx_{k+1})\bigr\|^2_{\bm\Sigma_{\text{ct}}(\Delta t_2)},
\end{equation}
where the residual $\bm\epsilon_{\text{ct}}$ is defined in Eq.~\eqref{eq:residual_ct}.
In Mode~A, $\tgtx\in\SE(3)$ for all keyframes; in Mode~B, $\tgtx$ may switch between $\SE(3)$ and $\R^3$ depending on the measurement type.
The same residual expression is instantiated on the appropriate manifold.
We model growth of prediction uncertainty with the propagation horizon via
\begin{equation}
    \bm\Sigma_{\text{ct}}(\Delta t_2) \;=\; s(\Delta t_2)\,\bar{\bm\Sigma}_{\text{ct}},
    \qquad\text{with } s(\Delta t_2)=\Delta t_2,
\end{equation}
using a linear scaling $s(\Delta t_2)$.

\paragraph{Measurement factors}
Measurement factors complement the motion prior. The USBL factor is
\begin{equation}
    \phi_{\text{usbl}}(\chx,\tgtx) \;=\;
    \bigl\|\, \bm h_{\text{usbl}}(\chx,\tgtx) - \bm z_{\text{usbl}} \,\bigr\|^2_{\bm\Sigma_{\text{usbl}}},
\end{equation}
with
\begin{equation}
    \bm h_{\text{usbl}}(\chx,\tgtx) = \bm t\!\bigl(\chxinv \tgtx\bigr)
    \label{eq:usbl_model_comp}
\end{equation}

Here $\bm t(\cdot)$ extracts the translation of an $\SE(3)$ element and
acts as the identity on $\R^3$ points.
When $\tgtx \in \SE(3)$, the expression in~\eqref{eq:usbl_model_comp} performs the
$\SE(3)$ group compose and then extracts translation. When $\tgtx \in \R^3$,
$\chxinv \tgtx$ is the $\SE(3)$ group action on a point,
so $\bm t(\cdot)$ simply returns that $3$-vector. In this case,
\eqref{eq:usbl_model_comp} evaluates to
\[
    \Rot{\Wf}{\Cf}^{\!\top}\bigl(\bm \tgtx - \bm t(\chx)\bigr)
\]

Optical and odometry factors are standard relative–pose constraints in
$\SE(3)$, differing only in whether they connect two chaser states or a
chaser–target pair.
\paragraph{Mode A roll–pitch prior}
To encode domain knowledge that the surface vessel remains upright, we add a
soft roll–pitch prior in Mode~A. Let $\Rot{\Wf}{\Tf}$ be the
orientation of the target and let $\psi(\Rot{\Wf}{\Tf})$ denote its yaw angle. Define
the yaw-aligned reference rotation $\bar R(\tgtx) \!=\! R_z\!\big(\psi(\Rot{\Wf}{\Tf})\big)$.
We penalize roll and pitch by measuring the rotation taking $\Rot{\Wf}{\Tf}$ to $\bar R(\tgtx)$ and projecting its logarithm onto the $(x,y)$ components (roll and pitch):
\begin{equation}
    \bm\epsilon_{\text{rp}}(\tgtx)
    \;=\;
    \mathbf{S}_{\text{rp}}\,
    \Log\!\big( {\Rot{\Wf}{\Tf}}^{\top}\,\bar R(\tgtx) \big),    \quad
    \mathbf{S}_{\text{rp}}
    \;=\;
    \begin{bmatrix} 1 & 0 & 0\\[0.3ex] 0 & 1 & 0 \end{bmatrix},
    \label{eq:rp_residual}
\end{equation}
where $\Log(\cdot)$ is the matrix logarithm on $\mathrm{SO}(3)$ yielding a
rotation-vector in $\mathbb{R}^3$. The corresponding factor is
\begin{equation}
    \phi_{\text{rp}}(\tgtx)
    \;=\;
    \bigl\|\,\bm\epsilon_{\text{rp}}(\tgtx)\,\bigr\|^2_{\bm\Sigma_{\text{rp}}}.
    \label{eq:rp_factor}
\end{equation}

\noindent\emph{Remarks.}
(i) This prior is not required for mathematical definition of the
estimation with the constant–twist factor; it is an application-specific,
soft constraint that rules out physically unrealistic inverted solutions
during USBL-only intervals.
(ii) In other domains (e.g., agile aerial targets), one can simply omit
$\phi_{\text{rp}}$ to keep all attitude DoFs free.
(iii) Jacobians follow by the chain rule through $\psi(\cdot)$, $R_z(\cdot)$,
and the $\mathrm{SO}(3)$ logarithm.

Keyframes are created whenever a new relative measurement arrives or when a fixed time interval (1~s) elapses without a measurement. In both modes, time-gated keyframes are still constrained by the constant-twist factor, ensuring that the trajectory remains defined during measurement gaps. In Mode~B, changes of representation $\R^3 \leftrightarrow \SE(3)$ at these boundaries are handled seamlessly by instantiating the residual in the appropriate manifold, following the boundary-factor framework of~\cite{espinoza2024boundaryfactors}.

The software developed for building and solving the factor graph is all based on the Georgia Tech Smoothing and Mapping (GTSAM) library~\cite{gtsam}.
\begin{figure*}[t]
    \centering
    \includegraphics[width=\textwidth]{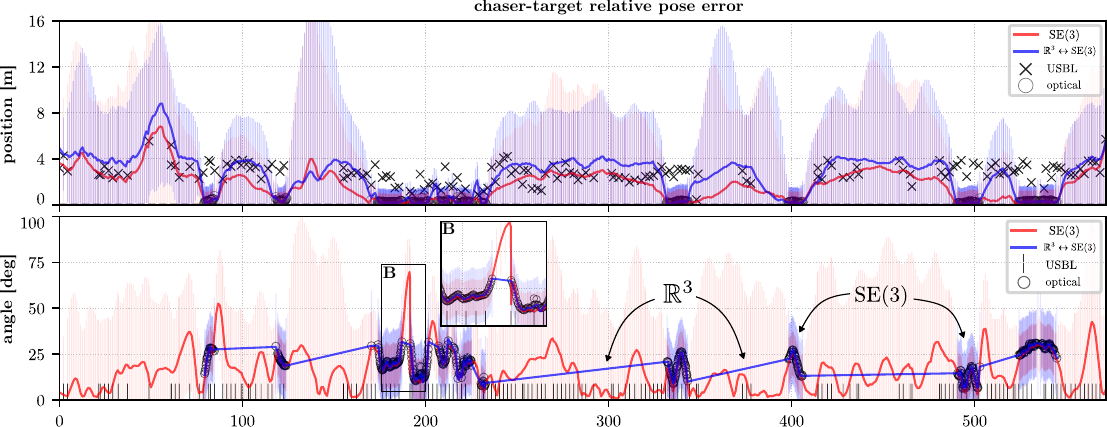}
    \caption{Mean (line) and $3\sigma$ bounds (shaded regions) for the relative position and angle errors of the estimated trajectories and raw measurements. The $\SE(3)$ and $\R^3$ annotations on the bottom plot point towards regions with specific sensor modalities.}
    \label{fig:relative_errors}
\end{figure*}

\begin{table}[]
    \caption{Means and standard deviations for the results in Fig.~\ref{fig:relative_errors}.}
    \label{tab:quantitative_results}
    \resizebox{\columnwidth}{!}{%
        \begin{tabular}{@{}cccc@{}}
            \toprule
            \multicolumn{1}{r}{Keyframe type:}                    & \textbf{USBL [m]} & \textbf{Optical [m]} & \textbf{All [m]}  \\ \midrule
            Mode A: $\mathrm{SE}(3)$--only                        & $1.524 \pm 1.182$ & $0.211 \pm 0.087$    & $1.187 \pm 1.272$ \\
            Mode B: $\mathbb{R}^3 \leftrightarrow \mathrm{SE}(3)$ & $2.458 \pm 1.616$ & $0.214 \pm 0.092$    & $1.757 \pm 1.778$ \\
            $\bm{z}_{\text{usbl}}$ (baseline)                     & $2.749 \pm 0.847$ & ---                  & ---               \\
            $\bm{t}(\bm{z}_{\text{opt}})$ (baseline)              & ---               & $0.218 \pm 0.102$    & ---               \\ \bottomrule
        \end{tabular}%
    }
\end{table}
\subsection{Results}
Figure~\ref{fig:full_trajectories} compares ground truth with fully smoothed
estimates for both deployment modes of the proposed prior:
\emph{Mode~A} ($\SE(3)$ only) and \emph{Mode~B}
($\R^3 \leftrightarrow \SE(3)$ via boundary factors).
The chaser trajectory is very accurate in both modes due to good initialization and
high-quality odometry. The target trajectory shows larger variation in USBL-only
segments, where orientation is unobserved and relative position measurements are
noisier. Both modes interpolate plausible target motion through extended gaps
with no measurements, which is significant because without a motion prior, the
$\SE(3)$ target state would be underconstrained during USBL-only intervals.
In Mode~A, the constant-twist prior preserves coupled rotational--translational
evolution on $\SE(3)$, producing arc-like extrapolations; in Mode~B, the state
switches to $\R^3$ and the prior reduces to constant translational velocity,
yielding straight-line extrapolations. Region~A in
Figure~\ref{fig:full_trajectories} highlights long gaps where the two modes
extrapolate differently but converge once optical resumes. This extrapolation prior to receiving the next measurement is highlighted in Figure~\ref{fig:moneyplot}.

Figure~\ref{fig:relative_errors} reports smoothed \emph{relative} pose errors.
The top panel shows position error. During optical keyframe intervals both modes agree
closely and track the optical measurements near ground truth. During USBL-only
and measurement-free intervals, error and uncertainty increase for both methods,
but Mode~A often lower-bounds the error of the raw USBL measurements and
is typically lower than Mode~B, consistent with the intuition that maintaining the
$\SE(3)$ representation can help regularize unobserved orientation while also
improving translation estimates. The bottom panel shows relative \emph{angle} error. For Mode~B, orientation is
only defined at optical keyframes while Mode~A maintains an orientation estimate
in between the bursts of optical measurements. In most optical keyframe windows, both modes follow the optical
pose accuracy. In longer gaps (e.g., Region~B), where both USBL and optical measurements are absent,
Mode~A continues to propagate the last observed angular trend from the optical
measurements. This can be beneficial if the target’s true motion is consistent
with that trend, or detrimental if the optical estimates just prior to the gap
already deviated from ground truth or if the target changes heading during the
gap, in which case angular error grows until the next measurement arrives.
Notably, even in these measurement-free intervals, the average angular error during gaps
between optical keyframes remains on par with that during the optical windows
themselves, suggesting that the constant-twist prior is able to maintain a
temporally consistent orientation trajectory rather than drifting arbitrarily.

To quantify these observations, Table~\ref{tab:quantitative_results} summarizes mean~$\pm$~std relative position
errors by keyframe type. In USBL-only windows, Mode~A outperforms Mode~B and the
raw USBL baseline (\(1.524\pm1.182\) vs.\ \(2.458\pm1.616\) and
\(2.749\pm0.847\)~m, respectively). In optical windows, the two modes are
similar---as expected given the much higher quality of the optical measurements.
Aggregated over all keyframes, Mode~A achieves \(\approx 1.5\times\) lower mean
position error than Mode~B (1.187 vs.\ 1.757~m), aligning with the qualitative
trends in Figs.~\ref{fig:full_trajectories} and~\ref{fig:relative_errors}.
(Orientation is not reported for Mode~B outside optical windows because the
target is represented in $\R^3$ there.)

\subsection{Discussion}
These results support the following two takeaways. First, the constant-twist prior enables a
solvable target trajectory in intervals that are otherwise underconstrained
and it can be instantiated either by keeping the state on $\SE(3)$ (Mode~A) or by switching to $\R^3$ when the measurements only provide relative position
(Mode~B), while preserving coupling between these transitions. Second, although Mode~A’s coupled twist can better stabilize translation during sparse USBL
intervals on average, it will propagate angular trends through gaps in the optical measurements. Our ternary implementation uses the most recent
two states to compute twist for the next state, which can make   extrapolation somewhat sensitive to
local noise.
\newpage
\section{Conclusion and Future Work}
\label{sec:conclusions}
During dynamic underwater proximity operations, the chasing agent must keep track of both its trajectory and its target's in order to execute a successful rendezvous. The problem with tracking a target underwater is the type of relative measurements available to the chaser for such scenarios (noisy, sparse, and asymmetric) underconstrain the trajectory estimation. To overcome this problem, we introduce a manifold-agnostic constant-twist motion prior that enforces temporally consistent estimates of the target's trajectory to overcome the challenging nature of the available measurements. We demonstrate the performance of the proposed motion prior using two different factor graph structures that instantiate the motion prior with two different representations depending on the nature of the desired target's state: either as a consistent pose in $\SE(3)$ or with a switching $\R^3 \leftrightarrow \SE(3)$ state. Both approaches demonstrate the ability to estimate the coupled trajectories of an AUV and a target in a real-world dynamic opto-acoustic scenario, enabling consistent estimation through under constrained USBL-only and fully constrained optical segments and improved relative tracking accuracy compared to the noisy measurements to the target. Although successful results were achieved, limitations regarding the specific implementation of the constant-twist priors in the factor graphs prevent online incremental operation required for real-time deployments. These future works include, but are not limited to: \emph{adjustable temporal horizons} (multi-horizon twist constraints) to trade aggressive trajectory estimation responsiveness for increased noise robustness; estimating target motion parameters from smoothed results to inform the prior; adding non-holonomic constraints directly on twist; and joint optimization of predicted target states and chaser controls to achieve a desired future relative pose.
\appendix
\section{Analytical Jacobians for Constant-Twist Factor}
\label{appendix:jacobians}

Let $T_{k-1}, T_k, T_{k+1} \in \mathcal{M}$, where $\mathcal{M}$ is a smooth manifold.
We use \emph{retraction} and \emph{local coordinates}
\[
    X \oplus \delta \,:\, \mathcal{M} \times T_X \mathcal{M} \to \mathcal{M},
    \qquad
    Y \ominus X \,:\, \mathcal{M} \times \mathcal{M} \to T_X \mathcal{M},
\]
with $\delta \in T_X \mathcal{M}$. Define the constant-twist prediction and residual:
\begin{align}
    \delta_1       & \triangleq T_k \ominus T_{k-1},                          \\
    \hat{\bm{\xi}} & \triangleq \frac{\delta_1}{\Delta t_1},                  \\
    \delta_2       & \triangleq \hat{\bm{\xi}} \cdot \Delta t_2,              \\
    \hat{T}_{k+1}  & \triangleq T_k \oplus \delta_2,                          \\
    e_k            & \triangleq T_{k+1} \ominus \hat{T}_{k+1}. \label{eq:e_k}
\end{align}

\subsection{Chain Rule Decomposition (General Manifold)}
\label{subsec:chain_rule_decomposition}
Applying the multivariable chain rule to \eqref{eq:e_k} gives
\begin{align}
    \frac{\partial e_k}{\partial T_{k-1}}
     & = J_{\hat{T}_{k+1}}^{e_k} \;
    J_{\delta_2}^{\hat{T}_{k+1}} \;
    J_{\hat{\bm{\xi}}}^{\delta_2} \;
    J_{\delta_1}^{\hat{\bm{\xi}}} \;
    J_{T_{k-1}}^{\delta_1},         \\[4pt]
    \frac{\partial e_k}{\partial T_k}
     & = \underbrace{
        J_{\hat{T}_{k+1}}^{e_k} \;
        J_{\delta_2}^{\hat{T}_{k+1}} \;
        J_{\hat{\bm{\xi}}}^{\delta_2} \;
        J_{\delta_1}^{\hat{\bm{\xi}}} \;
        J_{T_k}^{\delta_1}
    }_{\text{path via }\delta_1}
    \;+\;
    \underbrace{
        J_{\hat{T}_{k+1}}^{e_k} \;
        J_{T_k}^{\hat{T}_{k+1}}
    }_{\text{direct path}},         \\[4pt]
    \frac{\partial e_k}{\partial T_{k+1}}
     & = J_{T_{k+1}}^{e_k}.
\end{align}

Here $J_a^b := \partial b / \partial a$ denotes the Jacobian of $b$ with respect to $a$ on the state manifold~$\mathcal{M}$.

\subsection{Specialization to Lie Groups}
\label{subsec:specialization_to_lie_groups}
For a Lie group $\mathcal{M}$, we define the group operators
\[
    X \oplus \delta \triangleq X\,\Exp(\delta), \qquad
    Y \ominus X \triangleq \Log(X^{-1}Y),
\]
with the corresponding Jacobians (right-invariant local coordinates for $\ominus$, right-plus for $\oplus$):
\begin{align}
    J_X^{X \oplus \delta}
     & =
    \underbrace{\Ad_{\Exp(\delta)}^{-1}}_{\text{\cite[Eq.~(80)]{Sola2018MicroLie}}},
    \label{eq:jacobian-X-plus-delta}     \\[0.3em]
    J_\delta^{X \oplus \delta}
     & =
    \underbrace{J_r(\delta)}_{\text{\cite[Eq.~(81)]{Sola2018MicroLie}}},
    \label{eq:jacobian-delta-plus-delta} \\[0.3em]
    J_Y^{Y \ominus X}
     & =
    \underbrace{J_r^{-1}(Y \ominus X)}_{\text{\cite[Eq.~(83)]{Sola2018MicroLie}}},
    \label{eq:jacobian-Y-minus-X}        \\[0.3em]
    J_X^{Y \ominus X}
     & =
    \underbrace{-\,J_l^{-1}(Y \ominus X)}_{\text{\cite[Eq.~(82)]{Sola2018MicroLie}}},
    \label{eq:jacobian-X-minus-X}
\end{align}

where $J_l$ and $J_r$ are the left/right Jacobians on the Lie algebra, and $\Ad$ is the group adjoint.
Within the chain rule of the general case, the product $J_{\hat{\bm{\xi}}}^{\delta_2} J_{\delta_1}^{\hat{\bm{\xi}}}$ corresponds to converting
a displacement $\delta_1$ over $\Delta t_1$ into a velocity $\hat{\bm{\xi}}$,
and then back into a displacement $\delta_2$ over $\Delta t_2$.
Explicitly:
\begin{align}
    J_{\hat{\bm{\xi}}}^{\delta_2}
    \;J_{\delta_1}^{\hat{\bm{\xi}}}
     & =
    \underbrace{\frac{\partial \delta_2}{\partial \hat{\bm{\xi}}}}_{\Delta t_2 I_{d\times d}}
    \;
    \underbrace{\frac{\partial \hat{\bm{\xi}}}{\partial \delta_1}}_{\frac{1}{\Delta t_1} I_{d\times d}}
    \\[0.3em]
     & = \frac{\Delta t_2}{\Delta t_1}\, I_{d\times d}
    = \alpha\, I_{d\times d}.
    \label{eq:jacobian-alpha-I}
\end{align}

This scalar multiple of the identity will appear as \(\alpha I\) in the final Lie group Jacobians.

\subsection{Final Lie Group Jacobians for the Factor}
\label{subsec:final_lie_group_jacobians}
Plugging in the Jacobians from Section \ref{subsec:specialization_to_lie_groups} into the chain rule expressions from Section \ref{subsec:chain_rule_decomposition}, we obtain:
\begin{align}
    \frac{\partial e_k}{\partial T_{k-1}}
     & =
    \underbrace{\big(-J_l^{-1}(e_k)\big)}_{\text{$J_{\hat{T}_{k+1}}^{e_k}$ via \eqref{eq:jacobian-X-minus-X}}}
    \;\underbrace{J_r(\delta_2)}_{\text{$J_{\delta_2}^{\hat{T}_{k+1}}$ via \eqref{eq:jacobian-delta-plus-delta}}}
    \;\underbrace{\alpha I_{d\times d}}_{\text{$J_{\hat{\bm{\delta_1}}}^{\delta_2}$ via \eqref{eq:jacobian-alpha-I}}}
    \;\underbrace{\big(-J_l^{-1}(\delta_1)\big)}_{\text{$J_{T_{k-1}}^{\delta_1}$ via \eqref{eq:jacobian-X-minus-X}}},
    \label{eq:lie-d-ek-d-Tkm1} \\[4pt]
    \frac{\partial e_k}{\partial T_k}
     & =
    \underbrace{\big(-J_l^{-1}(e_k)\big)}_{\text{$J_{\hat{T}_{k+1}}^{e_k}$ via \eqref{eq:jacobian-X-minus-X}}}
    \;\underbrace{J_r(\delta_2)}_{\text{$J_{\delta_2}^{\hat{T}_{k+1}}$ via \eqref{eq:jacobian-delta-plus-delta}}}
    \;\underbrace{\alpha I_{d\times d}}_{\text{$J_{\hat{\bm{\delta_1}}}^{\delta_2}$ via \eqref{eq:jacobian-alpha-I}}}
    \;\underbrace{J_r^{-1}(\delta_1)}_{\text{$J_{T_k}^{\delta_1}$ via \eqref{eq:jacobian-Y-minus-X}}}
    \notag                     \\[-0.3em]
     & \quad+\;
    \underbrace{\big(-J_l^{-1}(e_k)\big)}_{\text{$J_{\hat{T}_{k+1}}^{e_k}$ via \eqref{eq:jacobian-X-minus-X}}}
    \;\underbrace{\Ad_{\Exp(\delta_2)}^{-1}}_{\text{$J_{T_k}^{\hat{T}_{k+1}}$ via \eqref{eq:jacobian-X-plus-delta}}},
    \label{eq:lie-d-ek-d-Tk}   \\[4pt]
    \frac{\partial e_k}{\partial T_{k+1}}
     & =
    \underbrace{J_r^{-1}(e_k)}_{\text{$J_{T_{k+1}}^{e_k}$ via \eqref{eq:jacobian-Y-minus-X}}}.
    \label{eq:lie-d-ek-d-Tkp1}
\end{align}
Equations \eqref{eq:lie-d-ek-d-Tkm1}–\eqref{eq:lie-d-ek-d-Tkp1} hold for any Lie group (with the chosen right-invariant local coordinates).

\subsection{SE(3) Case}
\label{subsec:se3_case}
If $\mathcal{M}=\SE(3)$, we use the standard adjoint
\[
    \Ad_T =
    \underbrace{\begin{bmatrix}
            R & [t]_\times R \\
            0 & R
        \end{bmatrix}}_{\text{\cite[Eq.~(175)]{Sola2018MicroLie}}},
    \qquad
    \Ad_T^{-1} =
    \underbrace{\begin{bmatrix}
            R^\top & -R^\top [t]_\times \\
            0      & R^\top
        \end{bmatrix}}_{\text{\cite[Eq.~(175)]{Sola2018MicroLie}}}.
\]

\paragraph{SE(3) Jacobian \emph{functions} required by the chain rule}
The ternary factor’s chain rule (Section~\ref{subsec:final_lie_group_jacobians}) calls the following
SE(3) Jacobian functions:
\begin{align}
     & \underbrace{J_r^{-1}(\delta)}_{\text{\cite[Eq.~(79)]{Sola2018MicroLie}}},\quad
    \underbrace{J_l^{-1}(\delta)}_{\text{\cite[Eq.~(82)]{Sola2018MicroLie}}},\quad
    \underbrace{\Ad_{\Exp(\delta)}^{-1}}_{\text{\cite[Eq.~(80)]{Sola2018MicroLie}}},\quad
    \underbrace{J_r(\delta)}_{\text{\cite[Eq.~(81)]{Sola2018MicroLie}}}.
\end{align}

\paragraph{Closed forms used by the blocks}
Write $\delta = [\,\bfrho;\,\bth\,] \in \mathbb{R}^6$ with translation $\bfrho$ and rotation $\bth$.
Note that this ordering places translation first; in contrast, GTSAM’s internal convention is
$\delta = [\,\bth;\,\bfrho\,]$ with rotation first.

\medskip
\noindent\emph{SO(3) left Jacobian inverse}
\begin{equation}
    \mjac{-1}{l}(\bth) =
    \underbrace{
        \mathbf I - \tfrac{1}{2}[\bth]_\times
        + \left( \tfrac{1}{\theta^2} - \tfrac{1 + \cos\theta}{2\theta\sin\theta} \right) [\bth]_\times^2
    }_{\text{\cite[Eq.~(146)]{Sola2018MicroLie}}}
    \label{eq:SO3LeftJacInv}
\end{equation}

\noindent\emph{SE(3) left Jacobian inverse (block form)}
\begin{equation}
    \mjac{-1}{l}(\bfrho,\bth) =
    \underbrace{
        \begin{bmatrix}
            \mjac{-1}{l}(\bth) & -\,\mjac{-1}{l}(\bth)\,\bfQ(\bfrho,\bth)\,\mjac{-1}{l}(\bth) \\
            \mathbf 0          & \mjac{-1}{l}(\bth)
        \end{bmatrix}
    }_{\text{\cite[Eq.~(179b)]{Sola2018MicroLie}}}
    \label{eq:SE3LeftJacInv}
\end{equation}

\noindent\emph{$Q(\bfrho,\bth)$ block in $J_l^{\SE(3)}$}
\begin{equation}
    \underbrace{
        \begin{aligned}
            \bfQ(\bfrho,\bth) ={} &
            \tfrac{1}{2}[\bfrho]_\times
            + \tfrac{\theta - \sin\theta}{\theta^3}                                                              \\
                                  & \quad \big( [\bth]_\times[\bfrho]_\times + [\bfrho]_\times[\bth]_\times
            + [\bth]_\times[\bfrho]_\times[\bth]_\times \big)                                                    \\[0.3em]
                                  & - \tfrac{1 - \tfrac{\theta^2}{2} - \cos\theta}{\theta^4}                     \\
                                  & \quad \big( [\bth]_\times^2[\bfrho]_\times + [\bfrho]_\times[\bth]_\times^2
            - 3[\bth]_\times[\bfrho]_\times[\bth]_\times \big)                                                   \\[0.3em]
                                  & - \tfrac{1}{2} \left( \tfrac{1 - \tfrac{\theta^2}{2} - \cos\theta}{\theta^4}
            - 3 \tfrac{\theta - \sin\theta - \tfrac{\theta^3}{6}}{\theta^5} \right)                              \\
                                  & \quad \times \big( [\bth]_\times[\bfrho]_\times[\bth]_\times^2
            + [\bth]_\times^2[\bfrho]_\times[\bth]_\times \big)
        \end{aligned}
    }_{\text{\cite[Eq.~(180)]{Sola2018MicroLie}}}
    \label{eq:Qmatrix_2}
\end{equation}

\noindent\emph{SE(3) right Jacobian inverse from the left Jacobian inverse}
\begin{equation}
    \mjac{-1}{r}(\bfrho,\bth) =
    \underbrace{\mjac{-1}{l}(-\bfrho,-\bth)}_{\text{\cite[Eq.~(76)]{Sola2018MicroLie}}}
    \label{eq:RightJacInvFromLeft}
\end{equation}

\paragraph{SE(3) left/right Jacobians (non-inverse) in block form}
Finally,
\begin{equation}
    \mjac{}{l}(\bfrho,\bth) =
    \underbrace{
        \begin{bmatrix}
            \mjac{}{l}(\bth) & \bfQ(\bfrho,\bth) \\
            \mathbf 0        & \mjac{}{l}(\bth)
        \end{bmatrix}
    }_{\text{\cite[Eq.~(179a)]{Sola2018MicroLie}}}
    \label{eq:SE3LeftJacobianBlock}
\end{equation}

\begin{equation}
    \mjac{}{r}(\bfrho,\bth) =
    \underbrace{\mjac{}{l}(-\bfrho,-\bth)}_{\text{\cite[Eq.~(76)]{Sola2018MicroLie}}}
    \label{eq:SE3RightJacobianFromLeft}
\end{equation}

\subsection{$\mathbb{R}^n$ Case}
If $\mathcal{M} = \mathbb{R}^n$ with the standard Euclidean structure,
the retraction and local coordinates reduce to ordinary vector addition and subtraction:
\[
    X \oplus \delta = X + \delta,
    \qquad
    Y \ominus X = Y - X.
\]
The Jacobians are constant:
\begin{align}
    J_Y^{\,Y \ominus X}          & = I_n,  &
    J_X^{\,Y \ominus X}          & = -I_n,   \\
    J_X^{\,X \oplus \delta}      & = I_n,  &
    J_\delta^{\,X \oplus \delta} & = I_n,
\end{align}
where $I_n$ is the $n \times n$ identity matrix.

For completeness (used by the constant–twist factor in the Lie group form), the Euclidean
adjoint and left/right Jacobians reduce to identities:
\begin{align}
  J_r^{-1}(\cdot) &= I_n, & J_l^{-1}(\cdot) &= I_n, \\
  J_r(\cdot)      &= I_n, & \Ad_{X}^{-1}     &= I_n,
\end{align}

Substituting these into the (Lie group) chain-rule expressions yields the Euclidean version
of the factor Jacobians.
\bibliography{references}

\begin{thebibliography}{10}
\providecommand{\url}[1]{#1}
\csname url@samestyle\endcsname
\providecommand{\newblock}{\relax}
\providecommand{\bibinfo}[2]{#2}
\providecommand{\BIBentrySTDinterwordspacing}{\spaceskip=0pt\relax}
\providecommand{\BIBentryALTinterwordstretchfactor}{4}
\providecommand{\BIBentryALTinterwordspacing}{\spaceskip=\fontdimen2\font plus
\BIBentryALTinterwordstretchfactor\fontdimen3\font minus
  \fontdimen4\font\relax}
\providecommand{\BIBforeignlanguage}[2]{{%
\expandafter\ifx\csname l@#1\endcsname\relax
\typeout{** WARNING: IEEEtran.bst: No hyphenation pattern has been}%
\typeout{** loaded for the language `#1'. Using the pattern for}%
\typeout{** the default language instead.}%
\else
\language=\csname l@#1\endcsname
\fi
#2}}
\providecommand{\BIBdecl}{\relax}
\BIBdecl

\bibitem{jacobi_inspection_2015}
\BIBentryALTinterwordspacing
M.~Jacobi, ``Autonomous inspection of underwater structures,'' \emph{Robotics
  and Autonomous Systems}, vol.~67, pp. 80--86, 2015, advances in Autonomous
  Underwater Robotics. [Online]. Available:
  \url{https://www.sciencedirect.com/science/article/pii/S0921889014002267}
\BIBentrySTDinterwordspacing

\bibitem{gonzalez_survey_2020}
\BIBentryALTinterwordspacing
J.~González-García, A.~Gómez-Espinosa, E.~Cuan-Urquizo, L.~G.
  García-Valdovinos, T.~Salgado-Jiménez, and J.~A.~E. Cabello, ``Autonomous
  underwater vehicles: Localization, navigation, and communication for
  collaborative missions,'' \emph{Applied Sciences}, vol.~10, no.~4, 2020.
  [Online]. Available: \url{https://www.mdpi.com/2076-3417/10/4/1256}
\BIBentrySTDinterwordspacing

\bibitem{teran_simultaneous_2023}
A.~Teran, A.~Teran, J.~Folkesson, N.~Rolleberg, P.~Sigray, and J.~Kuttenkeuler,
  ``Simultaneous trajectory estimation and mapping for autonomous underwater
  proximity operations,'' \emph{arXiv:2309.08780}, 2023.

\bibitem{dellaert_factor_2017}
\BIBentryALTinterwordspacing
F.~Dellaert and M.~Kaess, ``\BIBforeignlanguage{English}{Factor {Graphs} for
  {Robot} {Perception}},'' \emph{\BIBforeignlanguage{English}{Foundations and
  Trends® in Robotics}}, vol.~6, no. 1-2, pp. 1--139, Aug. 2017, publisher:
  Now Publishers, Inc. [Online]. Available:
  \url{https://www.nowpublishers.com/article/Details/ROB-043}
\BIBentrySTDinterwordspacing

\bibitem{gtsam}
\BIBentryALTinterwordspacing
F.~Dellaert and G.~Contributors, ``borglab/gtsam,'' May 2022. [Online].
  Available: \url{https://github.com/borglab/gtsam}
\BIBentrySTDinterwordspacing

\bibitem{Sola2018MicroLie}
\BIBentryALTinterwordspacing
J.~Sol{\`a}, J.~Deray, and D.~Atchuthan, ``A micro lie theory for state
  estimation in robotics,'' \emph{arXiv preprint arXiv:1812.01537}, 2018, last
  revised December 8, 2021. [Online]. Available:
  \url{https://arxiv.org/abs/1812.01537}
\BIBentrySTDinterwordspacing

\bibitem{espinoza2024boundaryfactors}
A.~T. Espinoza, A.~T. Espinoza, J.~Folkesson, P.~Sigray, and J.~Kuttenkeuler,
  ``Boundary factors for seamless state estimation between autonomous
  underwater docking phases,'' in \emph{2024 IEEE International Conference on
  Robotics and Automation (ICRA)}, 2024, pp. 9976--9982.

\bibitem{ruan_factor_2020}
L.~Ruan, S.~Chen, J.~Zhou, D.~Zeng, and Y.~Xu, ``A {Factor} {Graph} {Method}
  for {AUV} {Navigation} in the {Mobile} {Docking} {Progress},'' in
  \emph{Global {Oceans} 2020: {Singapore} – {U}.{S}. {Gulf} {Coast}}, Oct.
  2020, pp. 1--5, iSSN: 0197-7385.

\bibitem{yu_dynamic_2025}
\BIBentryALTinterwordspacing
T.~Yu, G.~Xu, Q.~Zhang, and T.~Liu, ``A dynamic docking system of auv based on
  bearings-only acoustic and visual coupled localization,'' \emph{Ocean
  Engineering}, vol. 328, p. 121001, 2025. [Online]. Available:
  \url{https://www.sciencedirect.com/science/article/pii/S0029801825007140}
\BIBentrySTDinterwordspacing

\bibitem{STEAM}
T.~D. Barfoot, C.~H. Tong, and S.~S{\"a}rkk{\"a}, ``Batch continuous-time
  trajectory estimation as exactly sparse gaussian process regression,'' in
  \emph{Proceedings of Robotics: Science and Systems (RSS)}, Berkeley, USA,
  12--16 July 2014.

\bibitem{Full_STEAM}
S.~Anderson and T.~D. Barfoot, ``Full steam ahead: Exactly sparse gaussian
  process regression for batch continuous-time trajectory estimation on
  se(3),'' in \emph{2015 IEEE/RSJ International Conference on Intelligent
  Robots and Systems (IROS)}, 2015, pp. 157--164.

\bibitem{Mukadam-IJRR-18}
M.~Mukadam, J.~Dong, X.~Yan, F.~Dellaert, and B.~Boots, ``Continuous-time
  {G}aussian process motion planning via probabilistic inference,'' vol.~37,
  no.~11, 2018, pp. 1319--1340.

\bibitem{dong2018sparse}
J.~Dong, M.~Mukadam, B.~Boots, and F.~Dellaert, ``Sparse {G}aussian processes
  on matrix {L}ie groups: A unified framework for optimizing continuous-time
  trajectories,'' in \emph{2018 IEEE International Conference on Robotics and
  Automation (ICRA)}.\hskip 1em plus 0.5em minus 0.4em\relax IEEE, 2018, pp.
  6497--6504.

\bibitem{petersen_integrated_2023}
M.~E. Petersen and R.~W. Beard, ``The integrated probabilistic data association
  filter adapted to lie groups,'' \emph{IEEE Transactions on Aerospace and
  Electronic Systems}, vol.~59, no.~3, pp. 2266--2285, 2023.

\bibitem{HERTZBERG201357}
\BIBentryALTinterwordspacing
C.~Hertzberg, R.~Wagner, U.~Frese, and L.~Schröder, ``Integrating generic
  sensor fusion algorithms with sound state representations through
  encapsulation of manifolds,'' \emph{Information Fusion}, vol.~14, no.~1, pp.
  57--77, 2013. [Online]. Available:
  \url{https://www.sciencedirect.com/science/article/pii/S1566253511000571}
\BIBentrySTDinterwordspacing

\bibitem{teran_consistent_2025}
A.~Terán~Espinoza, A.~Terán~Espinoza, C.~Deutsch, N.~Rolleberg, J.~Folkesson,
  P.~Sigray, and J.~Kuttenkeuler, ``A consistent dataset for dynamic underwater
  proximity operations,'' in \emph{OCEANS 2025 Brest}, 2025, pp. 01--09.

\end{thebibliography}

\end{document}